\documentclass[runningheads]{llncs}

\usepackage{accv}

\usepackage{accvabbrv}

\usepackage{graphicx}
\usepackage{enumitem}
\usepackage{booktabs}
\usepackage{multirow}
\usepackage{array}
\usepackage{pifont}

\usepackage[accsupp]{axessibility}  

\usepackage{hyperref}

\usepackage{orcidlink}

\begin{document}

\title{Improving Image Clustering with Artifacts Attenuation via Inference-Time Attention Engineering} 

\titlerunning{Inference-Time Attention Engineering}

\author{Kazumoto Nakamura\and
Yuji Nozawa\and
Yu-Chieh Lin\and \\
Kengo Nakata\and
Youyang Ng}

\authorrunning{K.~Nakamura et al.}

\institute{Kioxia Corporation \\
\email{\{kazumoto1.nakamura,yuji1.nozawa,yuchieh.lin,\\kengo1.nakata,youyang.ng\}@kioxia.com}}

\maketitle

\begin{abstract}
   The goal of this paper is to improve the performance of pretrained Vision Transformer (ViT) models, particularly DINOv2, in image clustering task without requiring re-training or fine-tuning. As model size increases, high-norm artifacts anomaly appears in the patches of multi-head attention. We observe that this anomaly leads to reduced accuracy in zero-shot image clustering. These artifacts are characterized by disproportionately large values in the attention map compared to other patch tokens. To address these artifacts, we propose an approach called Inference-Time Attention Engineering (ITAE), which manipulates attention function during inference. Specifically, we identify the artifacts by investigating one of the Query-Key-Value (QKV) patches in the multi-head attention and attenuate their corresponding attention values inside the pretrained models. ITAE shows improved clustering accuracy on multiple datasets by exhibiting more expressive features in latent space. Our findings highlight the potential of ITAE as a practical solution for reducing artifacts in pretrained ViT models and improving model performance in clustering tasks without the need for re-training or fine-tuning.
\end{abstract}

\section{Introduction}
\label{sec:intro}
The Transformer~\cite{vaswani2017attention} architecture revolutionized the field of natural language processing (NLP), and its success inspired the development of the Vision Transformer (ViT)~\cite{dosovitskiy2020image} for computer vision tasks. It has shown promising results across various downstream tasks~\cite{li2022exploring, ranftl2021vision, xie2021segformer, wang2021end} such as classification and retrieval. Various improvements to ViT have also been proposed~\cite{liu2021swin, yang2021focal, shen2023study}. However, ViT is known to require large amount of training data for it to achieve superior performance. In other words, large-scale training of ViT architecture potentially unlocks and scales up the ability of ViT models. DINO (self-distillation with no labels)~\cite{caron2021emerging}, a method of self-supervised learning that allows training without labels, offers a robust pretraining strategy to ViT that yields high recognition accuracy when fine-tuned. DINOv2~\cite{oquab2023dinov2} further extends the concept by pretraining the ViT model on a large scale visual dataset, offering significant improvements and robustness across various downstream tasks~\cite{Chen_2024_CVPR, Zou_2024_CVPR}. The introduction of state-of-the-art pretrained large vision models such as DINOv2 has led to the emergence of ``foundation models'', which challenges the longstanding practice of training small vision models for specific tasks. This shift could potentially upend the traditional approach to model development and deployment in computer vision. One critical task closely related to training strategy is unsupervised image clustering. This task involves analyzing datasets with unknown categories and labels. The training strategy plays a vital role in optimizing these unknown elements by extracting rich features for clustering but training-based image clusterings can hinder their applicability in time-sensitive scenarios. Our goal is to fully utilize the potential of pretrained ViT models for image clustering without requiring re-training or fine-tuning.

However, it was found that high-norm artifacts anomaly appears in the patches of the attention block in the multi-head attention module of ViT model trained with DINOv2~\cite{darcet2023vision}, limiting its potential. These artifacts are characterized by disproportionately large values in the attention map compared to other patch tokens. Such artifacts have been reported to exist in models with a large number of parameters. As the size of models increases, the issue of artifacts in ViT-based models becomes increasingly significant. In this study, we investigate these artifacts by paying our attention to the raw features in latent space generated by these ViT models pretrained by DINOv2. We consider the task of zero-shot image clustering, where unsupervised clustering algorithms such as K-Means are applied directly to the output features of a pretrained vision model. Unlike other tasks, zero-shot image clustering directly utilizes the features generated in latent-space, enabling us to directly observe the impact of artifacts on the resulting accuracy. In this paper, we define ``zero-shot image clustering'' as an unsupervised image clustering task by using pretrained models without in-domain training for a particular dataset. Note that a vision model pretrained on a large number of images is inevitably seeing similar images with the test datasets during its pretraining stage. Hence, this task is not a completely out-of-distribution clustering task and the definition of zero-shot in this case is limited to the aforementioned definition of clustering with a pretrained model.

In our study, we first identify the artifacts of the model by computing the \(L_2\) norms of one of the Query-Key-Value (QKV) patches within the attention block of the final layer. Through this process, we found that artifacts exist even in the smaller models, not limited to larger models as previously thought when identifying artifacts through the norms of output tokens~\cite{darcet2023vision}. We infer that by introducing measures to attenuate these artifacts, richer features could be generated across various model sizes. In prior study~\cite{darcet2023vision}, the role of artifacts was successfully moved from patch tokens to additional tokens called register tokens in the training process. This measure improved the performance of downstream tasks at the cost of model re-training. In this study, we aim to improve the image clustering performance of the pretrained model in a more efficient way. We formulate an approach to attenuate these artifacts that does not involve any re-training step.

To address these artifacts, in this paper, we propose an approach called Inference-Time Attention Engineering (ITAE), which manipulates attention function during inference. Specifically, we identify artifacts in the attention block of the model's final transformer layer by using QKV patches and attenuate the values of their corresponding attentions (\cref{fig:method}) during inference time. We further investigate the impact of artifacts anomaly on the zero-shot image clustering task, and employ our proposed approach in models where the artifacts are identified. We show that our proposed method improves clustering accuracy on multiple datasets. In addition, we show that our method is also effective and acting as a complimentary technique to models employing registers. This improvement is due to the more expressive nature of the model with ITAE. Our findings highlight the potential of ITAE as a practical solution for reducing artifacts in pretrained Vision Transformer models and improving model performance in clustering tasks without the need for re-training or fine-tuning.

Overall, our contributions are summarized as follow.
\begin{enumerate}[label=(\arabic*)]
   \item We identify the artifacts anomaly in DINOv2-based ViT models by computing the \(L_2\) norms of the QKV patches in the attention block of the multi-head attention module and evaluate baseline performance on zero-shot image clustering tasks.
   \item We formulate ITAE, a method to manipulate attention function during inference of ViT models to attenuate artifacts.
   \item We perform empirical study and show that our proposed method improved the zero-shot clustering accuracy on multiple datasets by eliminating the negative impact of artifacts on the clustering task for models with artifacts identified.
\end{enumerate}

\section{Related works}
\label{sec:related}

\subsection{Pretrained Large Vision Models and their Potential}
The development of pretrained vision models can be dated back particularly to a series of self-supervised learning strategy, especially those contrastive learning methods~\cite{chen2020simple, chen2020big, he2020momentum, chen2020improved, grill2020bootstrap, chen2021exploring}. These methods did not pretrain their models with ViT architecture, thus limiting their potential. However, since the emergence of ViT, there has been a growing interest in applying self-supervised and weakly-supervised learning to train ViT-based models, including DINO, CLIP~\cite{CLIP}, ALIGN~\cite{jia2021scaling} and MoCov3~\cite{chen2021empirical}. Unlike supervised learning, self-supervised learning offers the advantage of training with large amounts of unlabeled or weakly labeled data without the need for costly annotations. Notably, DINOv2 is a model pretrained on a substantial amount of data using DINO's approach. CLIP is a model pretrained on large amount of vision and language data pairs. Their robustness allows for easy adaptation to diverse downstream tasks. With only minimal fine-tuning and task-specific heads, these general pretrained models can be transformed to tackle specific tasks. However, some form of parametric learning is still necessary in this scenario. Here, we focus on a specific use case of pretrained vision models, where no fine-tuning whatsoever is required and the pretrained models are employed as-is~\cite{nara2024revisiting}. This concept is already widely seen in the field of NLP, where large language models (LLM) are used as-is in various NLP tasks through the introduction of prompt-engineering during inference time~\cite{gpt3, wei2022chain, kojima2024}. In prompt-engineering, only the input tokens are modified but not the LLM itself. For computer vision tasks such as image retrieval and k-NN, it is already possible to directly utilize the output features of pretrained model in calculating distance in latent space between data samples~\cite{nara2024revisiting}. We focus our study on another task that similarly able to utilize the output features of a pretrained model as-is but less studied in this way: unsupervised image clustering.

\subsection{Artifacts Anomaly in Pretrained ViT and Attention Engineering}
DINOv2 is a pretrained model that achieved state-of-the-art performance in various computer vision tasks. However, it has been noted that artifacts~\cite{darcet2023vision} in the form of patches with abnormally large norms can exist in ViT models such as DINOv2, DeiT III~\cite{touvron2021training}, and OpenCLIP~\cite{schuhmann2022laionb}. Previous study~\cite{darcet2023vision} addressed these artifacts by incorporating register tokens during the model training stage. In contrast, our approach applies inference-time attention engineering to the pretrained models. This allows us to mitigate the artifact issue without requiring model re-training. Several works on manipulating self-attention have been reported~\cite{yang2021focal, chen2023accumulated, zhai2023stabilizing, lee2021vision}. However, these manipulations occurred during the learning phase, whereas our work focuses solely on the implementation during inference time. The most relevant works to our study are SATA~\cite{chen2023accumulated} and LSA~\cite{lee2021vision}, which attempt to filter out unwanted attentions in ViT. However, these methods involve a learning phase and do not address the specific challenges of pretrained models and image clustering tasks. SATA used direct observation to suppress attention weights after the softmax function, while LSA introduced diagonal masking of attention and leveraged a temperature parameter to sharpen the attention values. In contrast, our approach attenuates the attention of artifact patches based on the observed artifacts in the QKV patches.

\subsection{Deep Image Clustering and Application of Pretrained Models}
For the image clustering task, there are two primary approaches to consider. The first approach~\cite{xie2016unsupervised, chang2017deep, caron2018deep, ji2019invariant, van2020scan, dilokthanakul2016deep, jiang2016variational, yang2019deep, yang2019deep_gaussian, mukherjee2019clustergan} focuses on developing learning methods that jointly optimize both image feature and clustering end-to-end. The second approach~\cite{hinton2006, wu2018unsupervised, tao2020clustering} is a two-stage method that involves feature extraction suitable for clustering, follow by applying clustering algorithms like K-Means for the actual clustering process. In the first approach, DAC~\cite{chang2017deep} formulates clustering as a binary pairwise-classification problem while DeepCluster~\cite{caron2018deep} repeatedly assigns pseudo-labels to learn the neural network. In the second approach, ID~\cite{wu2018unsupervised} formulates feature learning as a non-parametric instance discrimination problem. IDFD~\cite{tao2020clustering} proposes to perform instance discrimination and feature decorrelation simultaneously. In this work, we opt for the second approach that involves a two-stage setting, which affords better interpretability. Prior studies have employed in-domain training for each dataset, but this limitation can hinder their applicability in time-sensitive scenarios. In contrast, our proposed method leverages off-the-shelf pretrained vision models, offering a promising solution for image clustering tasks without the need for domain-specific training. However, the use of ViT-based pretrained models for image clustering has garnered limited attention, with only a handful of papers~\cite{zhou2022deep, adaloglou2023exploring, chu2023image, lowe} being published in this area. These works focused on improving the clustering phase of the algorithms while our method focuses on improving the feature extraction phase without fine-tuning. In this paper, we seek to harness the untapped potential of pretrained models through inference-time engineering techniques. Our work aims to illuminate the underexplored territory of image clustering using state-of-the-art pretrained large vision models.

\begin{figure}[t]
   \includegraphics[width=\textwidth,trim={0 2cm 0 1.5cm},clip]{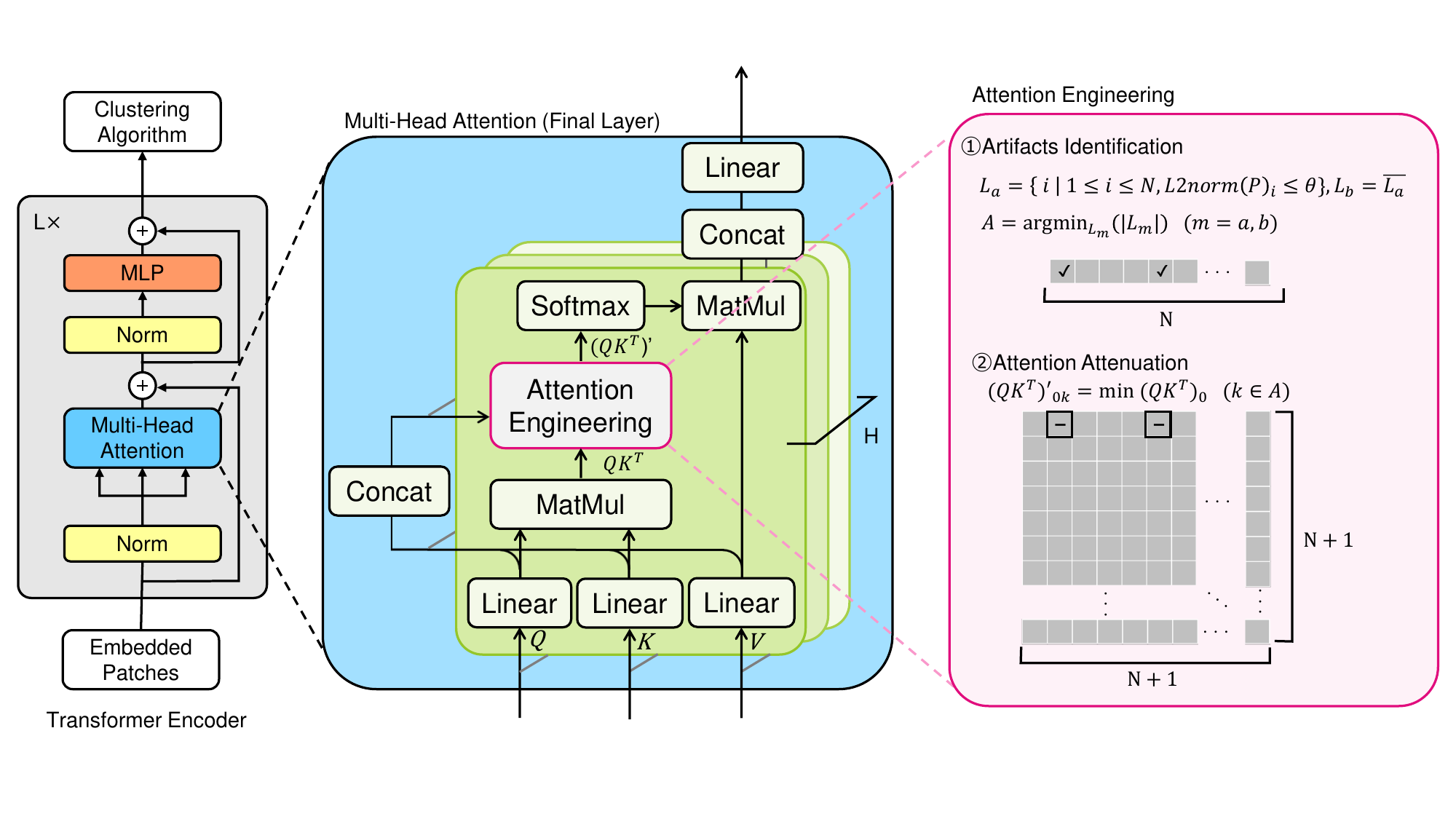}
   \caption{{\bf {Overall introduction of ITAE approach}}: We manipulate self-attention function of a pretrained vision model during inference. Specifically, we identify artifact in the final layer of the model's attention block during inference time by computing the \(L_2\) norms of the QKV patches, and attenuate the values of their corresponding attentions. \(N+1\) represents the length of the \(QK^T\) matrix, including the CLS token.}
   \label{fig:method}
\end{figure}

\section{Approach}
\label{sec:method}
In this section, we first revisit the mechanism of attention block in the multi-head attention module of transformer architecture. We then describe the approach we use to identify artifacts anomaly. Next, we explain the details of our proposed method of Inference-Time Attention Engineering (ITAE). In this paper, we define ``Attention Engineering'' as an attention manipulation technique that modifies the attention value of patches in attention block. The overall picture of our approach is shown in \cref{fig:method}. Finally, we describe the details of clustering algorithm (K-Means) that we apply in this study but note that the investigation of clustering algorithm itself is out of the scope of this paper.

\subsection{Self-Attention Block in Vision Transformer}
In ViT, attention in the Self-Attention (SA) block is calculated from the three matrices query, key, value as follows~\cite{dosovitskiy2020image}.
\begin{align}
   \label{eq:sa}
   \text{{SA}}(Q, K, V) = \text{{Attention}}(Q, K)V  = \text{{softmax}}\left(\frac{{QK^T}}{{\sqrt{d_k}}}\right)V,
\end{align}
where \(Q\) denotes the query matrix, \(K\) denotes the key matrix, and \(V\) denotes the value matrix. Also, \(d_k\) denotes the number of dimensions of the key.

\subsection{Identifying the Artifacts in Self-Attention Block}
Previous study~\cite{darcet2023vision} identified artifacts based on the feature norms of the output tokens of the model. However, we argue that artifacts are better analyzed within the self-attention block. We utilize the \(L_2\) norms distribution of the QKV patches, \(P\) within the self-attention block across all heads in the final layer. Using QKV is also computationally efficient as no back pass loop from the output is needed. In our study, we select query among the QKV patches to identify artifacts as we observed that norms distribution of query patches shows clearer bimodality between normal patches and artifacts. Note that it is also possible to use key or value patches to identify artifacts as they differ only by a linear projection. From this point onwards, we will use the term QKV to refer to query patches. In our work, the \(L_2\) norms distribution is collected across a dataset. We divide the \(L_2\) norms into two groups \(L_a, L_b \in \lbrace 1, 2, ..., N \rbrace\) with a threshold \(\theta\) and identify those patches in the minority group as artifacts, \(A\), where
\begin{align}
   L_a =& \:\lbrace i\:| 1 \leq i \leq N, \:L2norm(P)_i \leq \theta \rbrace, \:L_b = \overline{L_a},\\
   \label{eq:argmin}
   A =& \:\operatorname*{arg min}_{L_m}(|L_m|) \:\:\:\:\:(m = a, b),
\end{align}
and where
\begin{align}
   {L2norm(P)_i} = \left(\sum_{1\le h\le H} \sum_{1\le j\le d_p} (\frac{{P^h_{ij}}}{{\sqrt{d_p}}})^2\right)^{\frac{1}{2}}.
\end{align}
Here, \(H\) represents the number of heads in the multi-head attention module, \(N\) represents the number of patches and \(d_p\) denotes the number of dimensions of \(P\). Through our observation, the optimal value of \(\theta\) is roughly dataset-agnostic and model-dependent. This observation is consistent with the previous study~\cite{darcet2023vision} that the characteristic of artifacts depends on the scale of the models. Hence, we only need to pre-determine the value of \(\theta\). \(\theta\) is not modified for different datasets in our study. Our approach of analyzing artifacts via the QKV patches in self-attention block allows us to identify artifacts present in smaller ViT models that are difficult to recognize in the feature norms of the output tokens of the model. Specific analysis of \(L_2\) norms of QKV patches will be discussed in detail in \cref{subsec:art}.
 
\subsection{Inference-Time Attention Engineering}
After identifying the artifacts in self-attention block, we apply attention engineering by manipulating attention function during inference. Specifically, we attenuate the values of the identified artifact patches in the self-attention block of the model's final transformer layer. For every identified artifact, \(k\) in \(A\), the attention value of the artifact is transformed into a minimum value independently for every head in the multi-head attention module by the equation
\begin{align}
   ({QK^T})'_{ik} = \min_{1{\leq}j{\leq}N}\left(({QK^T})_{ij}\right) \: \: \: \: \: \left( k\in A\right),
\end{align}
where \(i\) and \(j\) have the same dimension. However, since we only focus on the final layer of the model and we use only the output CLS token when applying clustering algorithm, only the row \(0\) that corresponds to the CLS token needs to be calculated in our proposed approach. Consequently, we apply the following part of attention function in our study:
\begin{align}
   \label{eq:ITAE}
   ({QK^T})'_{0k} = \min_{1{\leq}j{\leq}N}\left(({QK^T})_{0j}\right) \: \: \: \: \: \left( k\in A\right). 
\end{align}
The full self-attention function can then be calculated through \cref{eq:sa} with component from \cref{eq:ITAE}. In our experiment, we also apply the same function in \cref{eq:ITAE} to models pretrained with register tokens. The pretrained models with register tokens we utilize in this study have a fixed number of register tokens, which is 4. As mentioned above, we extract the output features of CLS tokens in latent space for the subsequent clustering tasks. Our focus on applying the proposed method to the final layer of the model is justified by the simplicity of its modification, and the observability of the effect, where the output features are directly applied to clustering algorithm. 

\subsection{Clustering Algorithm}
Finally, we discuss the clustering algorithm that we utilize in our study. Note that in this work, our focus is on the improvement of feature extractor and the latent features it produces towards the performance of image clustering through ITAE. Tuning of clustering algorithm is out of the scope of this study. Since we do not modify the format of the latent features, any feasible clustering algorithm could be applied with our approach. Classical clustering algorithms like K-Means, spectral clustering and agglomerative clustering are among the candidates. Modern learnable dense algorithms are also applicable. In our work, we apply K-Means, one of the simplest and robust clustering algorithms available, and show in the subsequent section that our proposed method is effective in creating a richer feature space for image clustering task.

\section{Experiments}
\label{sec:expe}

\subsection{Experimental Settings}
In this study, we evaluate the performance of the zero-shot image clustering task on four common clustering datasets: Tiny ImageNet~\cite{tinyimagenet}, CIFAR-100~\cite{cifar}, CIFAR-10~\cite{cifar} and STL-10~\cite{coates2011analysis}. Tiny ImageNet contains image data of 200 classes extracted from the ImageNet dataset~\cite{ILSVRC15} and downsized to 64$\times$64 colored images. Each class has 500 training images, 50 validation images, and 50 test images. The validation images were used for the experiments. CIFAR-10 is a dataset consisting of 10 classes (5,000 training images and 1,000 test images for each class) with an image size of 32$\times$32 pixels. For the experiments, all 10,000 test images were used. CIFAR-100 is a dataset consisting of 100 classes (500 training images and 100 test images for each class) with an image size of 32$\times$32 pixels. All 10,000 test images were used for the experiments. STL-10 is also a dataset derived from ImageNet dataset, with image size 96$\times$96 pixels. It consists of 10 classes of data, 100,000 unlabeled data, 500 training images for each class, and 800 test images for each class. The test image set of 8,000 images was used for the experiments. For all datasets, the size of the images was standardized by resizing them to 224$\times$224 pixels before feeding them into the pretrained model to perform the evaluation. This preprocessing step ensured uniformity of the input dimensions for all datasets.

For ViT models, we use the publicly available models pretrained by DINOv2 in our experiments (\url{https://github.com/facebookresearch/dinov2/tree/main}). Specifically, we use the small (ViT-S/14 distilled), base (ViT-B/14 distilled), large (ViT-L/14 distilled) and giant (ViT-g/14) variation of ViT models pretrained on DINOv2 for our experiments with models without register tokens. Similarly, we use the same small, base, large and giant variation of ViT models but with register tokens pretrained on DINOv2 for our experiments with models with register tokens. Note that the small, base and large models are distilled from the giant models.

For the clustering algorithm, we use the standard K-Means algorithm without any prior parameters fine-tuning. 
We extracted the feature representation of each image by normalizing the CLS token output from the final layer of the model. As the initialization of K-Means can impact its results, for each experimental setting, we conducted 20 sets of 25 clustering runs (a total of 500 runs) to account for this initial value dependence. We calculate the mean and standard error of accuracy across these multiple clustering sets and report their values in this paper. Note that we do not fine-tune our models for clustering hence we report zero-shot image clustering results. As stated in \cref{sec:intro}, we define ``zero-shot image clustering'' as an unsupervised image clustering task by using pretrained models without in-domain training for a particular dataset.

For quantitative comparison, we compare the clustering accuracy of our proposed method with the original DINOv2 without applying ITAE. DINOv2 by itself achieved state-of-the-art performance across various downstream tasks. Similarly, for image clustering with pretrained model, DINOv2 is one of the most competitive models available today and this model presents us a strong baseline for comparison. We also compare our method to previous work where the model is pretrained with register tokens~\cite{darcet2023vision}. We use standard clustering accuracy (ACC), normalized mutual information (NMI), and adjusted rand index (ARI) as the metrics to measure clustering performance.

\begin{figure}[t]
   \begin{center}
      \begin{tabular}{cc}
         \fbox{\includegraphics[width=5.8cm]{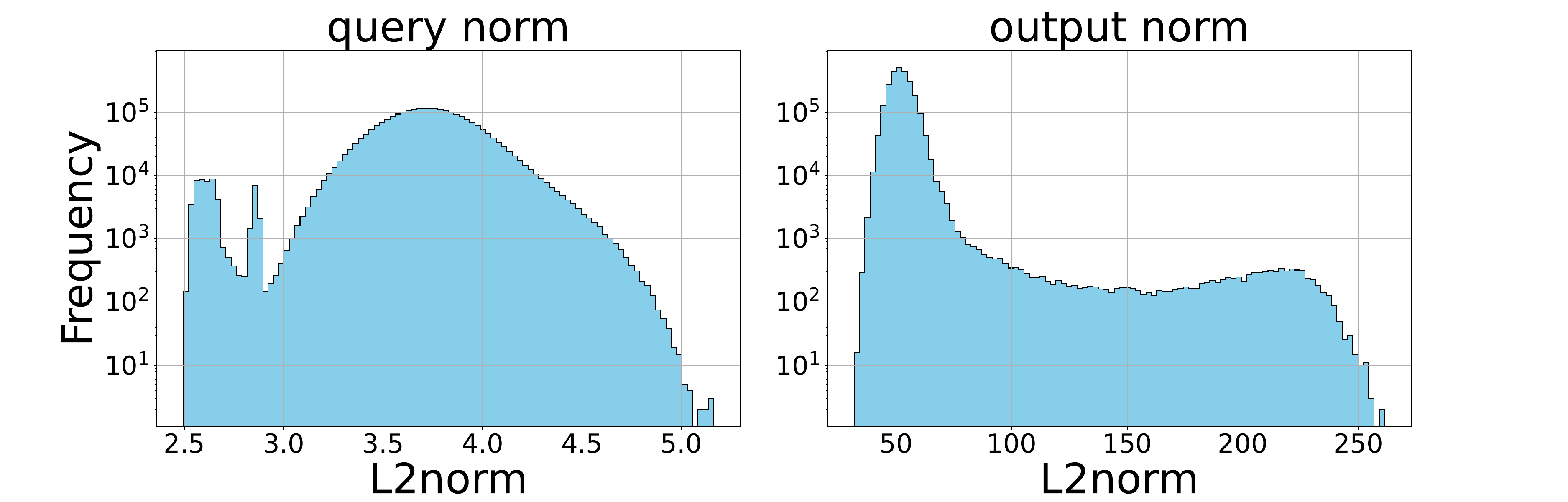}}&
         \fbox{\includegraphics[width=5.8cm]{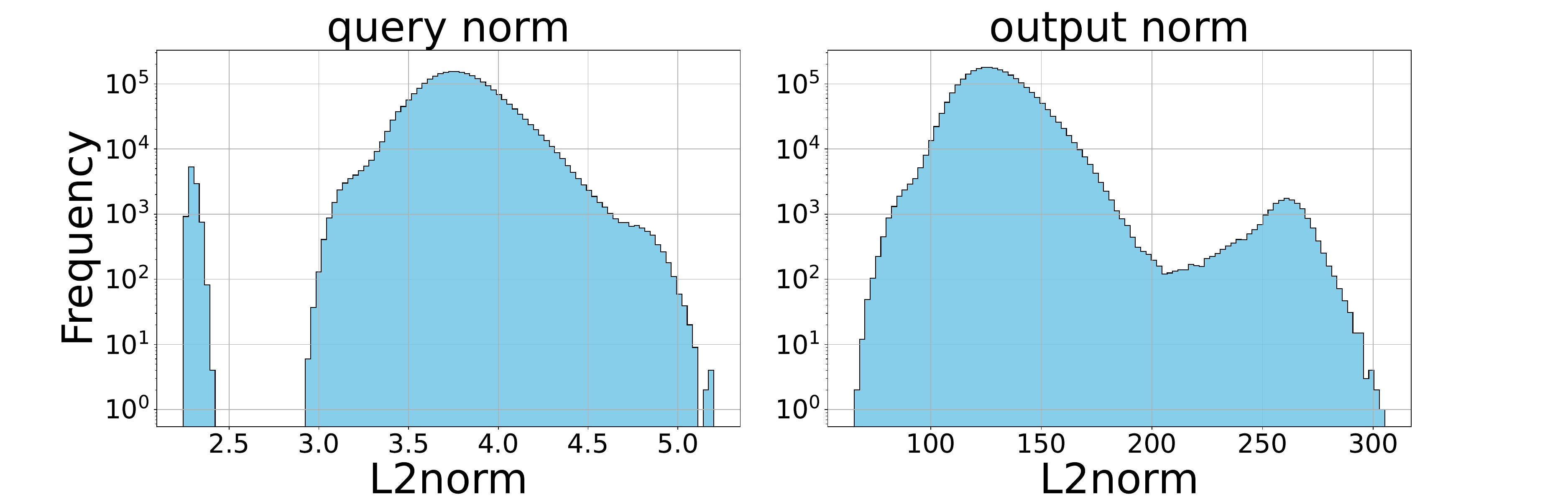}}\\
         without register&with register
      \end{tabular}
   \end{center}
   \caption{The left plots show the \(L_2\) norms distribution for model pretrained without register tokens while the right plots show the \(L_2\) norms distribution including register tokens for model pretrained with register tokens. \textit{Query norm} graphs show the \(L_2\) norms distribution of QKV patches excluding CLS token. \textit{Output norm} graphs show the \(L_2\) norms distribution of the final output. Clear bimodality is observed in the \(L_2\) norms distribution of QKV patches (model: ViT-B/14 distilled, dataset: CIFAR-100).}
      \label{fig:hist}
\end{figure}
   
\subsection{Artifact Identification}
\label{subsec:art}
A distribution analysis of the \(L_2\) norms of the QKV patches in self-attention block and the \(L_2\) norms of the output of the final layer of the Base model is presented in \cref{fig:hist}. Previous study has identified anomalous tokens based on the \(L_2\) norms distribution of the final output. For the model pretrained without register tokens, the \(L_2\) norms distribution of the patch tokens in the final output does not exhibit clear bimodality. Interestingly, the \(L_2\) norms distribution of the QKV patches does show bimodality, suggesting that the presence of artifacts can be determined using a threshold value in the \(L_2\) norms distribution of the QKV patches.  On the other hand, for the model pretrained with the register tokens, the \(L_2\) norms distribution of the patch tokens alone does not display bimodality~\cite{darcet2023vision}. By including both the patch tokens and the register tokens in the distribution analysis, it is confirmed that a portion of the register tokens functions as an artifact, as depicted in \cref{fig:hist}. Additionally, a comparison between models pretrained with and without register tokens reveals that the model incorporating the register tokens exhibits more pronounced bimodality. Through visually inspecting the bimodality in the \(L_2\) norms distribution of the QKV patches, we can easily set the value of $\theta$ for each model and identify the minority group as artifacts. Note that the value of $\theta$ is predicted to be roughly dataset-agnostic so we fix the value of $\theta$ throughout our experiments unless stated otherwise. Specifically, we set $\theta$ to 2.0 for small model pretrained with register tokens through visual inspection. For other models pretrained with and without register tokens, we set $\theta$ to a common value of 3.0.

\begin{table}[t]
   \caption{Image clustering results. \textit{original} denotes the original DINOv2 model while \textit{registers} denotes DINOv2 model pretrained with register tokens as implemented in previous work~\cite{darcet2023vision}. Our proposed method outperformed the baseline in all cases and previous work in most cases (model: ViT-B/14 distilled).}
   \begin{center}
   \resizebox{\columnwidth}{!}{
   \begin{tabular}{|c|c|c|c|c|c|c|}
   \hline
   Method & Metric & CIFAR-10 & CIFAR-100 & STL-10 &Tiny ImageNet\\
   \hline\hline
   \multirow{3}{*}{original}  & ACC & $83.63\pm 1.13$ & $64.26\pm 0.30$& $75.65\pm 1.04$ & $67.81\pm 0.24$ \\
                              & ARI & $77.75\pm 1.08$ & $48.94\pm 0.18$& $61.49\pm 1.57$  & $50.68\pm 0.25$ \\
                              & NMI & $86.04\pm 0.46$ & $76.58\pm 0.09$& $81.61\pm 0.72$  & $81.28\pm 0.06$ \\
   \hline
   \multirow{3}{*}{registers} & ACC & $82.12\pm 1.35$ & $\mathbf{66.79\pm 0.23}$ & $72.70\pm 1.13$ & $\mathbf{68.88\pm 0.21}$ \\
                             & ARI & $73.50\pm 1.25$ & $50.30\pm 0.21$  & $56.92\pm 1.37$ & $50.13\pm 0.23$ \\
                             & NMI & $84.70\pm 0.41$ & $\mathbf{78.42\pm 0.08}$ & $79.47\pm 0.54$ & $81.69\pm 0.05$ \\
   \hline
   \multirow{3}{*}{ours} & ACC &  $\mathbf{84.49\pm 1.19}$ & $65.02\pm 0.14$ & $\mathbf{82.76\pm 1.27}$ & $68.23\pm 0.25$ \\
                         & ARI &  $\mathbf{79.46\pm 1.14}$ & $\mathbf{50.53\pm 0.18}$ & $\mathbf{75.94\pm 1.60}$ &  $\mathbf{52.27\pm 0.22}$  \\
                         & NMI &  $\mathbf{86.82\pm 0.51}$ & $77.10\pm 0.08$  & $\mathbf{88.18\pm 0.62}$ & $\mathbf{81.78\pm 0.07}$ \\
   \hline
   \end{tabular}
   }
   \end{center}
   \label{tab:clu_base}
\end{table}

\subsection{Image Clustering Result}
The results of the proposed method, previous work and the baseline are presented in \cref{tab:clu_base}. Notably, by applying our method, consistent improvements in performance are observed. The most substantial enhancement in accuracy is observed in the STL-10 dataset, with a remarkable increase of 7.11 for ACC, 14.46 for ARI, and 6.57 for NMI. Overall, an average accuracy improvement of 2.24 was observed for our method compared to baseline. We also observe that accuracy of baseline is lower than previous work in 2 datasets, indicating the negative impact brought by the artifacts. However, interestingly, model pretrained with register tokens are not always performing better when comparing to the original baseline model. In fact, the accuracy on the datasets is reduced by -0.27 in average. On the other hand, our method, without any re-training, outperformed previous work in most cases. We speculate that models pretrained with register tokens managed to reduce the negative effect brought by the artifacts during pretraining but sacrificed global optimization of the model to a certain extent.
 
   \begin{table}[t]
      \caption{Image clustering result across various model sizes (\textit{small}: ViT-S/14 distilled, \textit{base}: ViT-B/14 distilled, \textit{large}: ViT-L/14 distilled, \textit{giant}: ViT-g/14) reported in ACC. For model size \textit{small}, we did not apply our method due to the lack of bimodality confirmation. Our proposed method outperformed the strong baseline in all cases across model sizes. Interestingly, models pretrained with register tokens are not always performing better when comparing to the original models.}
      \centering
      \resizebox{\columnwidth}{!}{
      \begin{tabular}{|c|c|c|c|c|c|}
          \hline
          Dataset & Model Size & original & registers & ours & registers + ours \\
                 \cline{3-6}
          \hline
          \multirow{4}{*}{CIFAR-10}       & small & $70.92\pm 1.54$ & $77.11\pm 1.60$  &        -                & $\mathbf{81.57\pm1.04}$ \\
                                          & base  & $83.63\pm 1.13$ & $82.12\pm 1.35$ & $\mathbf{84.49\pm 1.19}$ & $        82.61\pm1.40$  \\
                                          & large & $82.16\pm 1.48$ & $78.67\pm 1.50$ & $\mathbf{82.49\pm 1.55}$ & $        79.92\pm1.47$  \\
                                          & giant & $78.09\pm 1.25$ & $76.38\pm 1.44$ & $\mathbf{78.59\pm 1.91}$ & $        77.42\pm1.69$  \\
          \hline
          \multirow{4}{*}{CIFAR-100}      & small & $50.84\pm 0.24$ & $57.33\pm 0.20$ &        -                 & $\mathbf{60.80\pm0.27}$  \\
                                          & base  & $64.26\pm 0.30$ & $66.79\pm 0.23$ & $        65.02\pm 0.14$  & $\mathbf{68.85\pm0.28}$ \\
                                          & large & $68.69\pm 0.34$ & $68.01\pm 0.38$ & $\mathbf{69.04\pm 0.22}$ & $        68.98\pm0.31$ \\
                                          & giant & $68.99\pm 0.39$ & $69.22\pm 0.27$ & $        69.50\pm 0.28$  & $\mathbf{69.94\pm0.25}$ \\
          \hline
          \multirow{4}{*}{STL-10}         & small & $83.33\pm 1.69$ & $85.20\pm 2.04$ &        -                 & $\mathbf{85.89\pm1.45}$ \\
                                          & base  & $75.65\pm 1.04$ & $72.70\pm 1.13$ & $\mathbf{82.76\pm 1.27}$ & $        78.71\pm1.14$ \\
                                          & large & $65.78\pm 1.22$ & $56.84\pm 1.21$ & $\mathbf{70.51\pm 1.42}$ & $        59.62\pm1.49$ \\
                                          & giant & $55.91\pm 1.14$ & $53.73\pm 1.75$ & $\mathbf{56.01\pm 0.93}$ & $        55.06\pm1.41$ \\
          \hline
          \multirow{4}{*}{Tiny ImageNet}  & small & $55.49\pm 0.19$ & $57.78\pm 0.15$ &        -                 & $\mathbf{59.56\pm0.14}$ \\
                                          & base  & $67.81\pm 0.24$ & $68.88\pm 0.21$ & $        68.23\pm 0.25 $ & $\mathbf{69.62\pm0.18}$ \\
                                          & large & $71.98\pm 0.15$ & $71.53\pm 0.19$ & $\mathbf{73.19\pm 0.21}$ & $        72.53\pm0.17$ \\
                                          & giant & $73.25\pm 0.16$ & $73.44\pm 0.17$ & $        73.54\pm 0.17 $ & $\mathbf{73.80\pm0.20}$ \\
          \hline
      \end{tabular}
      }
      \label{tab:clu_models}
  \end{table}

\subsection{Analysis with Different Model Sizes}
\Cref{tab:clu_models} summarizes the findings regarding the relationship between model size and accuracy. It is observed that the optimal model size for achieving the highest accuracy varies depending on the dataset. Specifically, for the Tiny ImageNet dataset, accuracy consistently improves as the model size increases. Conversely, for the STL-10 dataset, accuracy tends to decrease as the model size increases. These results show that the optimal model size for image clustering is dataset-dependent. In addition to exploring the impact of model size, we also evaluated the effectiveness of our method on both original models and models pretrained with register tokens. In this case, we apply our method to models pretrained with register tokens, effectively attenuating those attentions absorbed by register tokens. We show this result in the ``registers + ours'' column of \cref{tab:clu_models}. Our experiments demonstrated similar improvement in accuracy for both types of models. Specifically, with the exception of the ViT-S/14 distilled, our method improved the accuracy by an average of 1.43 for the original model and 1.56 for the model pretrained with register tokens. This indicates that our method is robust and can enhance the clustering performance regardless of the presence or absence of register components. When observing across model sizes, similar to results in \cref{tab:clu_base}, models pretrained with register tokens are not always performing better when comparing to models pretrained without register tokens.

\section{Discussion}
\label{sec:disc}

\begin{figure}[h!]
   \centering
   \begin{tabular}{cc}
      \includegraphics[width=5.5cm]{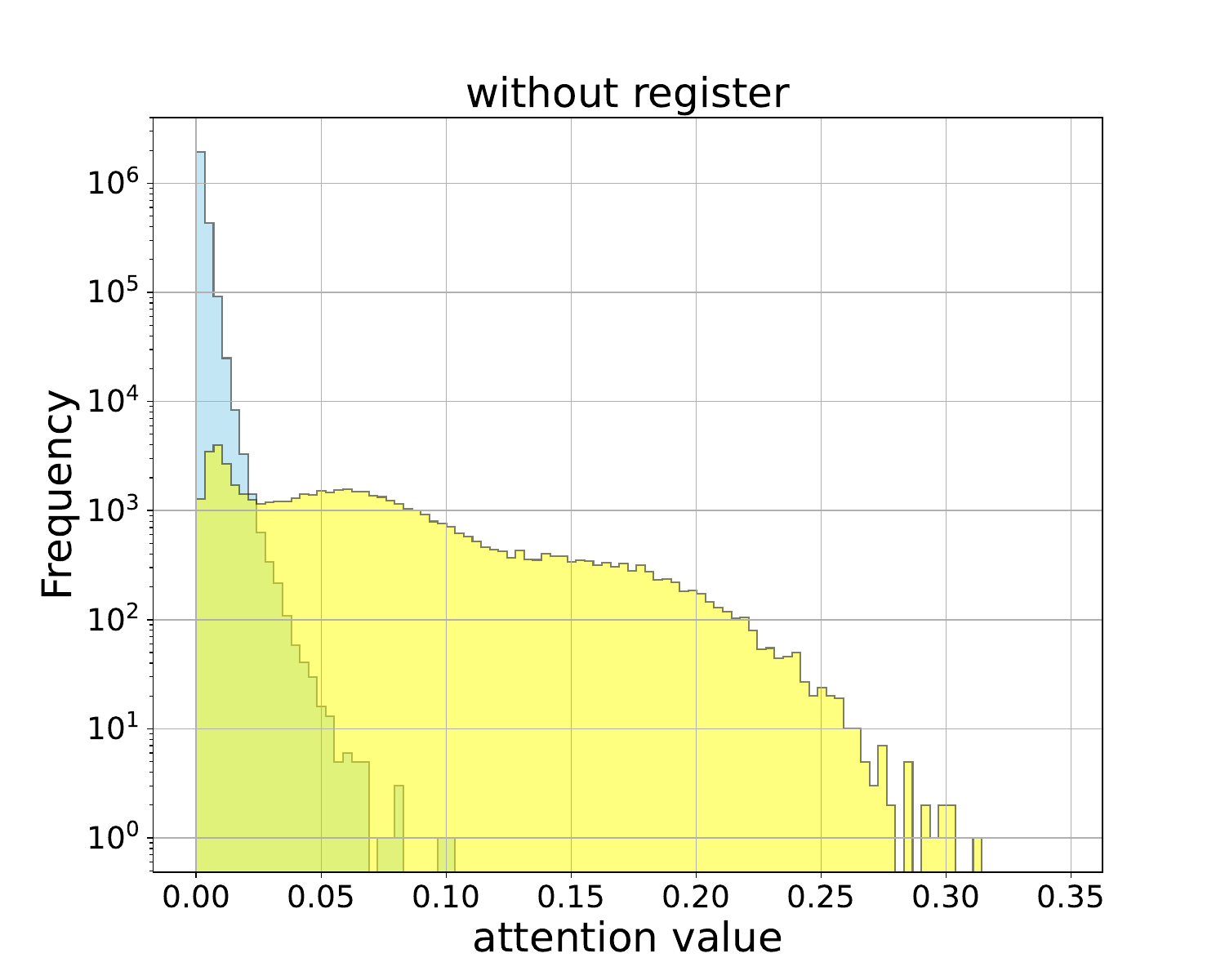}&
      \includegraphics[width=5.5cm]{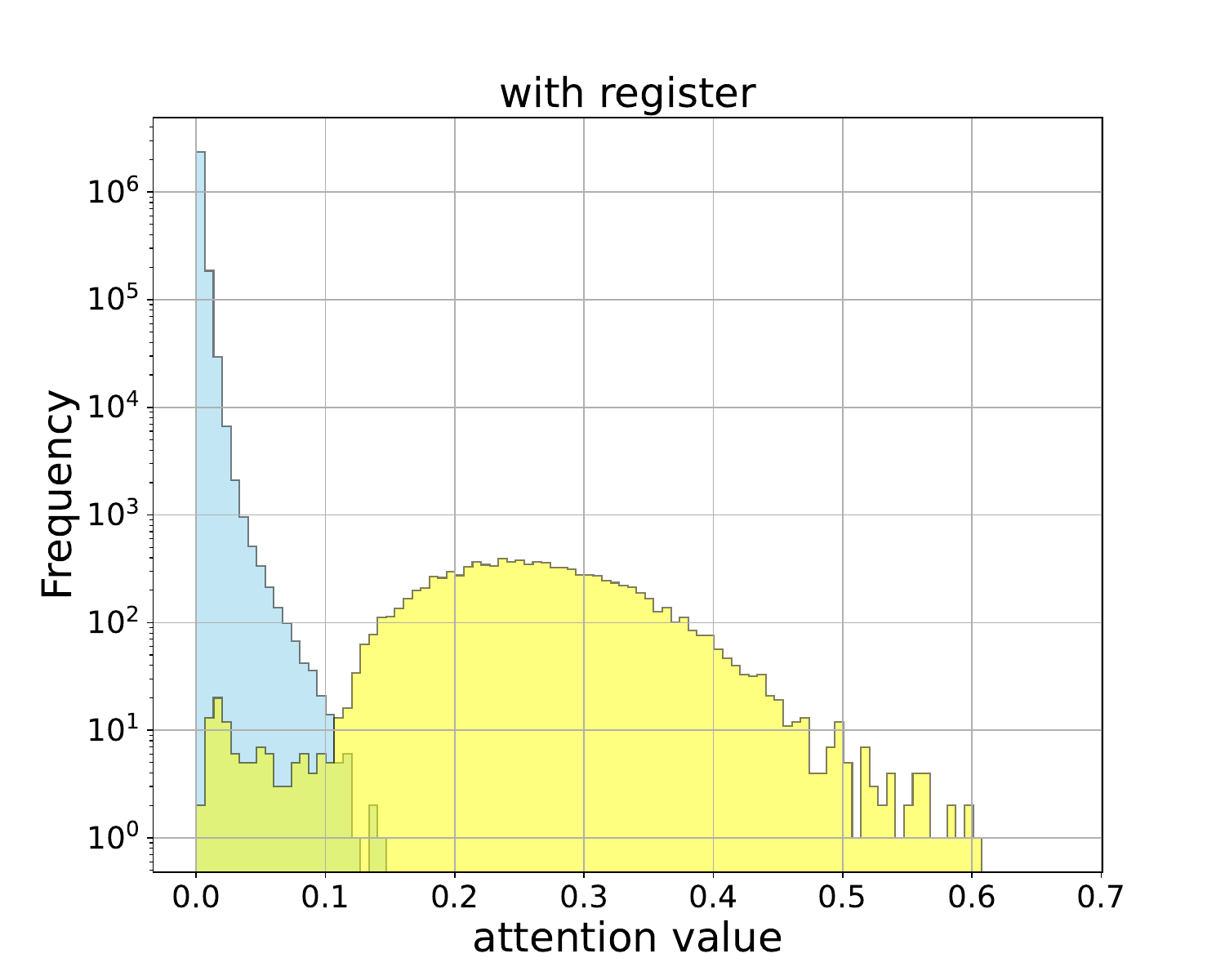}\\
   \end{tabular}
   \caption{Histogram of attention values: Yellow colored distribution shows the artifact patches identified while blue colored distribution shows normal patches. The left and right plots are for model pretrained without register tokens and model pretrained with register tokens, respectively (model: ViT-B/14 distilled, dataset: CIFAR-100). Best viewed in color. }
   \label{fig:attn_hist}
\end{figure}

\begin{figure}[h!]
   \centering
   \begin{tabular}{cc}
      \includegraphics[width=5.8cm]{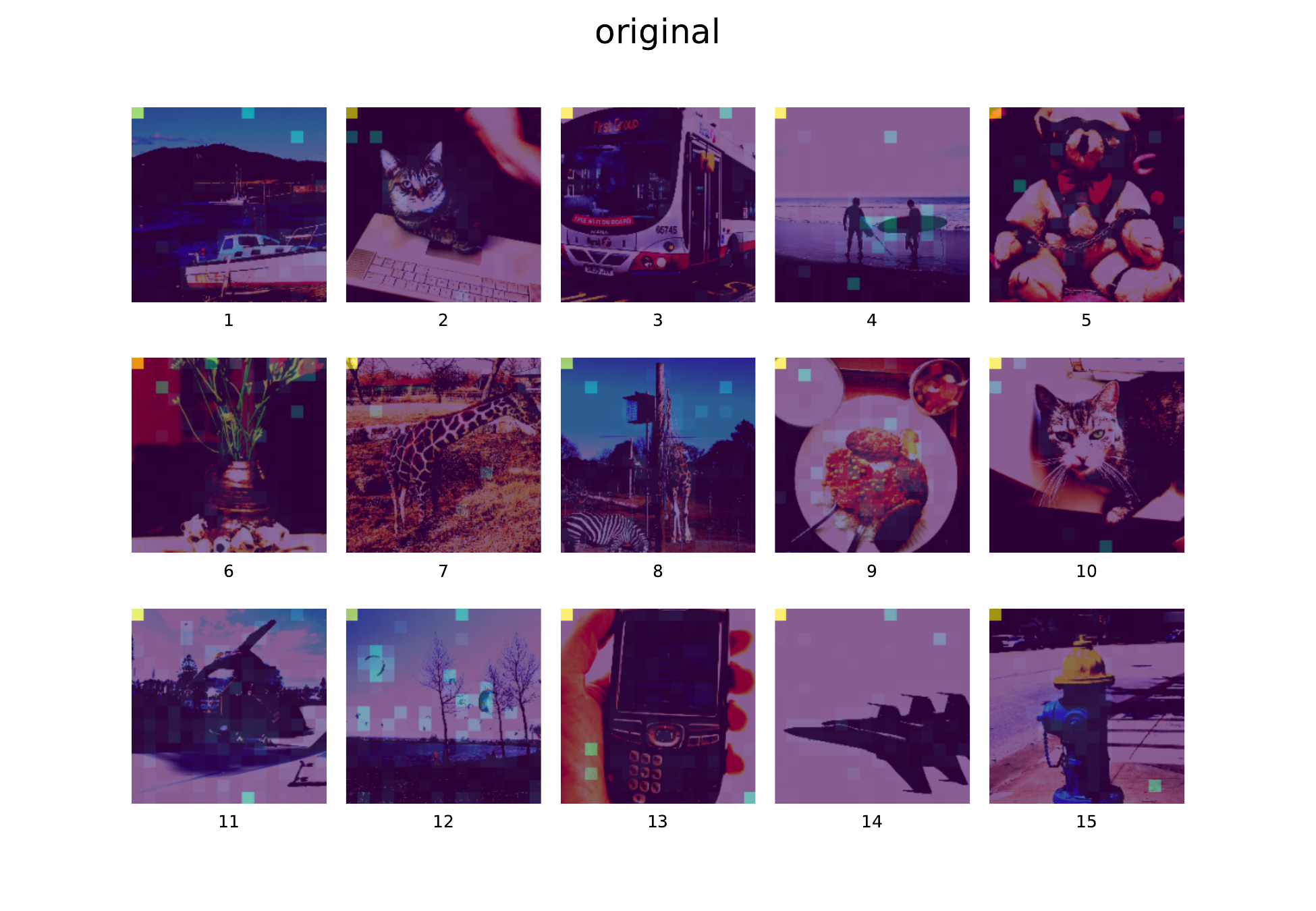}&
      \includegraphics[width=5.8cm]{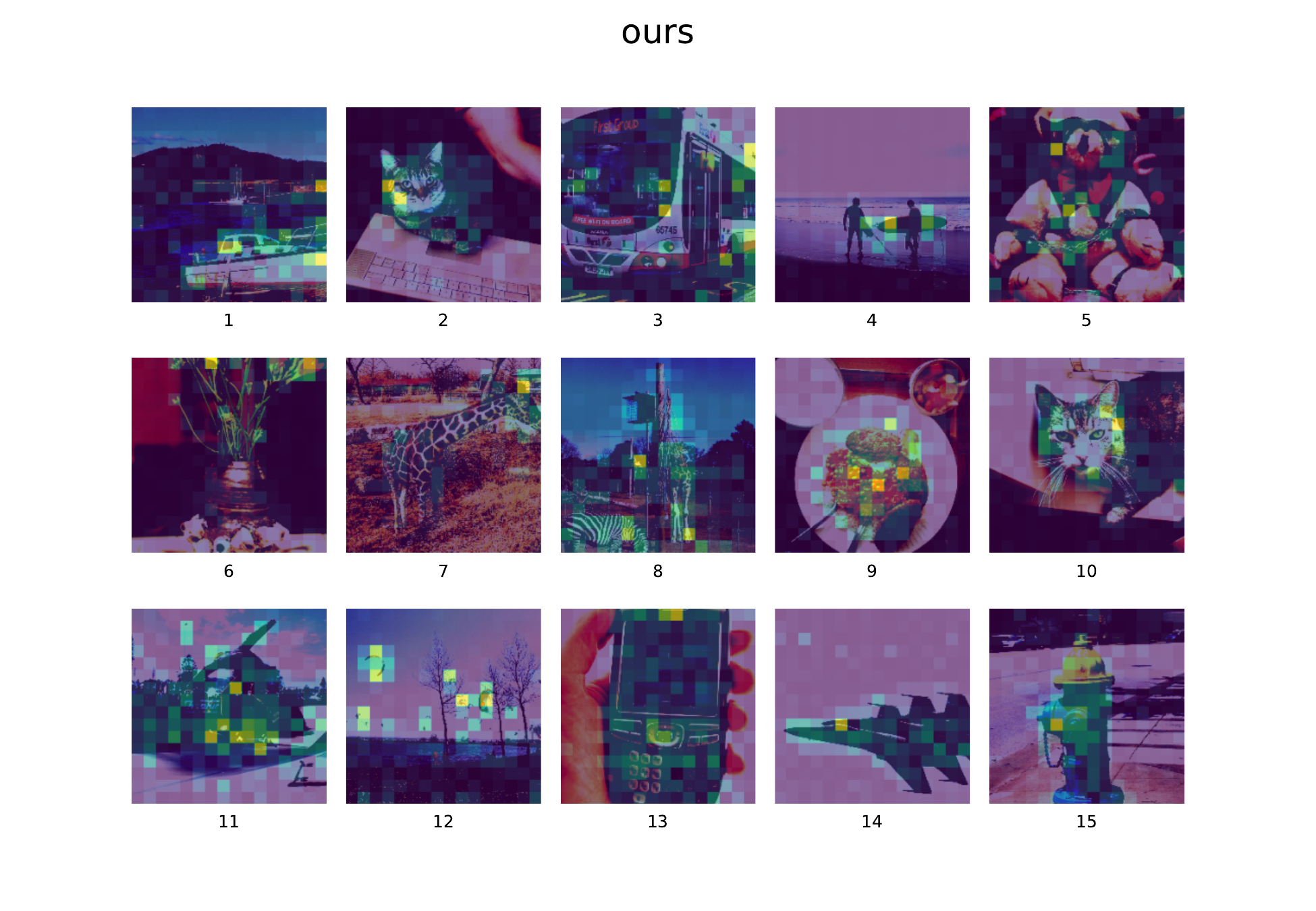}
   \end{tabular}
   \caption{Attention map: The left maps are the attention maps of original model and the right maps are the attention maps of the model incorporating our proposed method. The attention map is the averaged map across each attention head in the multi-head attention module. Best viewed in color. Details and licenses for the images are provided in supplementary material (model: ViT-B/14 distilled, dataset: MS COCO). }
   \label{fig:attention}
\end{figure}

\subsection{Attention Values and Maps Visualization}
A histogram analysis of the attention values associated with the identified artifacts in our proposed method is shown in \cref{fig:attn_hist}. These attention values are collected from the $\text{{Attention}}(Q,K)_{0j}  ~\text{{for}}~ 1\leq j \leq N$ in \cref{eq:sa}. Yellow colored distribution shows the artifact patches identified while blue colored distribution shows normal patches. The distribution shows that the identified artifact's attention values are substantially larger than normal. This observation suggests that the artifact's attention values contribute significantly to the output features, effectively masking the function of normal token features. By attenuating these identified artifacts during inference time, our proposed method enhances the model's expressiveness by redistributing attentions from the artifact to other tokens.
In addition, the impact of our proposed method on the attention map is shown in \cref{fig:attention}. The left maps are the attention maps of original model and the right maps are the attention maps of the model incorporating our proposed method. We show images from the dataset of MS COCO~\cite{coco} for these visualization due to its clear license info. We observe that our proposed method succeeds in expanding the attentions that were previously concentrated in certain areas of the attention map without any re-training or fine-tuning. These richer attentions are beneficial to tasks that directly employing output feature representations of the model, particular the zero-shot image clustering task studied in this paper.

\begin{figure}[t]
   \centering
   \includegraphics[width=0.5\textwidth]{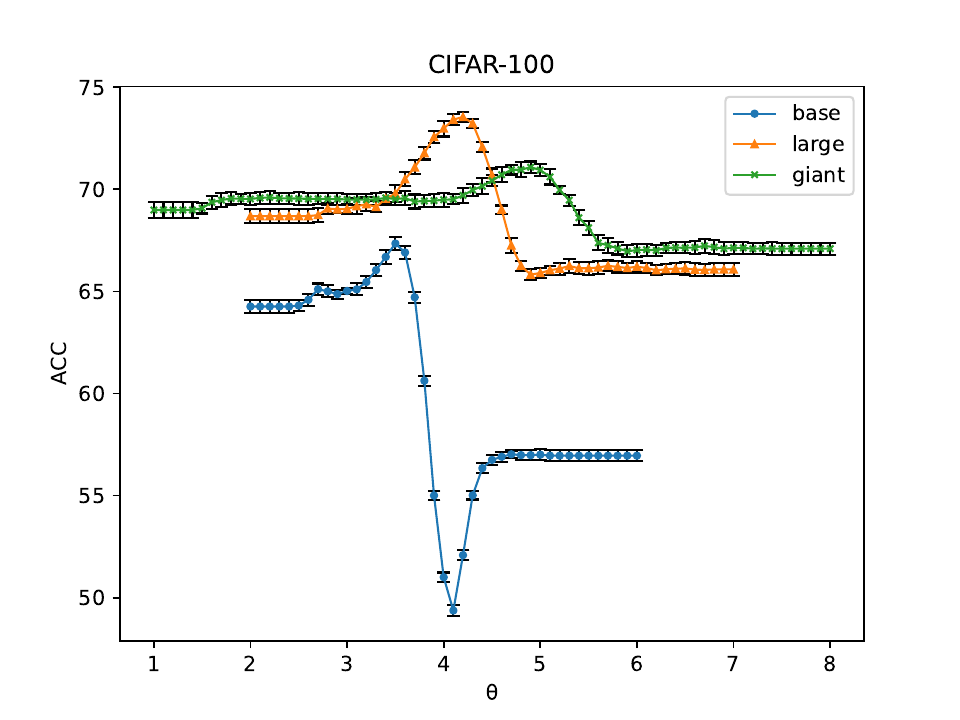}
   \caption{$\theta$-accuracy graph: Change in clustering accuracy (ACC) when $\theta$ is varied from 1.0 to 8.0 for each model (base: ViT-B/14 distilled, large: ViT-L/14 distilled, giant: ViT-g/14, dataset: CIFAR-100). Best viewed in color. }
   \label{fig:threshold}
\end{figure}

\subsection{Ablation Study of Threshold Values for Identifying Artifacts}
\label{subsec:thr}
In this session, we present an ablation study to investigate the impact of the threshold $\theta$ on clustering accuracy. To conduct this study, we adjust the threshold within the range of QKV patches. We remove \cref{eq:argmin} and treat \(L_a\) as artifacts \(A\) directly to perform a parameter scan. The results are illustrated in \cref{fig:threshold}, which displays the change in clustering accuracy corresponding to different threshold values for each model. In the following discussion, we will mainly consider the base model but note that similar observations can be made for other models.

Initially, at the lowest threshold, the clustering accuracy is similar to that of the original model. This is because tokens are not ignored at this stage. As we gradually increase the threshold, specifically around $\theta = 2.5$, the accuracy starts to improve. This improvement can be attributed to the exclusion of artifacts, which are now taken into consideration for accuracy calculation. Continuing our analysis, we observed that the accuracy continues to increase as the threshold reaches approximately $\theta = 3.5$. This is due to the removal of artifacts from the clustering process. However, beyond this threshold, we noticed a decline in accuracy. We hypothesize that this decline is a result of the attenuation of tokens required for clustering and the increased presence of non-artifact tokens that are irrelevant for clustering. As the threshold value $\theta$ increases, so does the value of $\text{{Attention}}(Q, K)_{00}$. This leads to the attention mechanism utilizing the CLS tokens as they are. In the base model, around the threshold of $\theta = 4.5$, most tokens other than the CLS tokens are attenuated, resulting in convergence of accuracy. At this point, there is a significant improvement in accuracy compared to the lowest accuracy value in the base model. However, it is worth noting that the degree of improvement varies across different models.

An intriguing finding from our study is that the accuracy peaks in the range greater than $\theta = 3$, where artifacts have already been ignored. This suggests that, in addition to artifacts, there are tokens that were previously considered normal but are not necessary for clustering. By appropriately processing these tokens, it is possible to further enhance the clustering accuracy.

\subsection{Generalization of the Proposed Method to Other Tasks}

\begin{table}[t]
   \caption{k-NN classification results across various model sizes (\textit{base}: ViT-B/14 distilled, \textit{large}: ViT-L/14 distilled, \textit{giant}: ViT-g/14) reported in accuracy.}
   \centering
   \resizebox{\columnwidth}{!}{
   \begin{tabular}{|c|c|c|c|c|c|}
       \hline
       Dataset & Model Size &  original & registers & ours & registers + ours \\
              \cline{3-6}
       \hline
       \multirow{3}{*}{ImageNet-1k}    & base  & $82.04$ & $82.02$ & $82.07$ & $\mathbf{82.35}$ \\
                                       & large & $83.50$ & $83.84$ & $83.62$ & $\mathbf{83.87}$ \\
                                       & giant & $83.51$ & $83.65$ & $83.54$ & $\mathbf{83.72}$ \\
       \hline
       \multirow{3}{*}{CIFAR-100}      & base  & $87.31$ & $87.60$ & $87.58$ & $\mathbf{88.09}$ \\
                                       & large & $91.12$ & $90.88$ & $\mathbf{91.39}$ & $91.01$ \\
                                       & giant & $91.79$ & $91.59$ & $\mathbf{91.99}$ & $91.81$ \\
       \hline
   \end{tabular}
   }
   \label{tab:tasks}
\end{table}

To evaluate the tasks generalizability of our proposed method, we conduct experiments with k-NN classification task~\cite{nakata2022}. In our experiment, we follow standard k-NN implementation and evaluation metric in Ref.~\cite{oquab2023dinov2} and set the value of $k$ to $10$. For datasets, since large scale datasets are applicable to k-NN classification, we use the full set of ImageNet-1k~\cite{ILSVRC15}, as well as CIFAR-100. The results are shown in \cref{tab:tasks}. It is observed that our method consistently improves over original model by a slight margin for different model sizes. Previous work of \textit{registers}~\cite{darcet2023vision} managed to improve the accuracy by a bigger margin for \textit{large} \& \textit{giant (ImageNet-1k)}, and \textit{base (CIFAR-100)} but performance degradation occurred for others. The complementary aspect of \textit{registers} and our method is extensively visible here as \textit{registers + ours} achieves highest accuracy in more categories. We report other ablation studies in the supplementary material.

\section{Conclusion}
In conclusion, our study has first re-identified the presence of artifacts in models of smaller size which were previously believed to be artifact-free. We then successfully improved the zero-shot image clustering accuracy by addressing the negative impact of these artifacts. Specifically, we achieved this by introducing ITAE, which manipulates attention function during inference. Our findings highlight the potential of ITAE as a practical solution for reducing artifacts in pretrained Vision Transformer models and improving model performance in clustering tasks without the need for re-training or fine-tuning. Potential future works include optimizing attention allocation to further improve model performance.


%
%
\bibliographystyle{splncs04}
\bibliography{mainbib}

\clearpage

\newcommand\beginsupplement{%
         \setcounter{equation}{0}
         \renewcommand{\theequation}{A\arabic{equation}}
         \setcounter{figure}{0}
         \renewcommand{\thefigure}{A\arabic{figure}}
         \setcounter{table}{0}
         \renewcommand{\thetable}{A\arabic{table}}
         \setcounter{section}{0}
         \renewcommand{\thesection}{A\arabic{section}}
         \setcounter{chapter}{0}
         \renewcommand{\thechapter}{A\arabic{chapter}}
}

\beginsupplement

\chapter*{Supplementary Material}

\section{Extended Ablation Studies}

\subsection{Generalization of the Proposed Method to Other Models with Different Pretraining Paradigms}

To evaluate the models generalizability of our proposed method towards different pretraining paradigms, we conduct additional experiments using weakly-supervised pretrained models of CLIP~\cite{CLIP} and supervised pretrained models of DeiT III~\cite{touvron2021training}. We follow the same experimental protocols outlined in our paper. We use publicly available pretrained models in our experiments (\textit{CLIP base\footnote{\url{https://openaipublic.azureedge.net/clip/models/5806e77cd80f8b59890b7e101eabd078d9fb84e6937f9e85e4ecb61988df416f/ViT-B-16.pt}}, CLIP large\footnote{\url{https://openaipublic.azureedge.net/clip/models/b8cca3fd41ae0c99ba7e8951adf17d267cdb84cd88be6f7c2e0eca1737a03836/ViT-L-14.pt}}, DeiT III base\footnote{\url{https://dl.fbaipublicfiles.com/deit/deit_3_base_224_21k.pth}}, DeiT III large\footnote{\url{https://dl.fbaipublicfiles.com/deit/deit_3_large_224_21k.pth}}}). We set $\theta$ to 2.0 for all CLIP models, 2.7 for DeiT III base model and 3.25 for DeiT III large model. Note that since DeiT III was trained for ImageNet-1k, to preserve zero-shot generalization discussion, we select only CIFAR-10 and CIFAR-100 for evaluation in the case of DeiT III. The results of both clustering and k-NN classification are shown in \cref{tab:models}. For CLIP models, the performance improves mostly with our method, similar to the results observed with DINOv2. For DeiT III, there is no obvious performance improvement, which aligns with the findings from the linear evaluation in Table 2 of Ref.~\cite{darcet2023vision}. Ref.~\cite{darcet2023vision} stated that pretraining paradigm seems to play a role in the characteristics of artifacts as CLIP and DeiT-III show artifacts at sizes smaller than DINOv2. We speculate that the supervised nature of DeiT III overfits the models to a particular dataset, decreasing the potential of performance extension during inference-time attention manipulation. We further analyzed the \(L_2\) norms distribution in CLIP and DeiT-III models and found that ITAE successfully identified and attenuated artifacts, similar to DINOv2. We did not test \textit{registers} and \textit{registers + ours} settings as pretrained CLIP and DeiT III models with registers are not publicly available. However, we speculate that the complementary synergy between our method and \textit{registers}~\cite{darcet2023vision} will enhance these models.

\begin{table}[t]
   \caption{Clustering \& k-NN classification results across various pretraining paradigms and model sizes (\textit{DINOv2 base}: ViT-B/14 distilled, \textit{DINOv2 large}: ViT-L/14 distilled, \textit{CLIP base}: ViT-B/16, \textit{CLIP large}: ViT-L/14, \textit{DeiT III base}: ViT-B/16, \textit{DeiT III large}: ViT-L/16). Clustering results are reported in ACC while k-NN classification results are reported in standard k-NN classification accuracy.}
   \centering
   \resizebox{1.0\columnwidth}{!}{
   \begin{tabular}{|c|c|c|c|c|c|}
      \hline
      Dataset & Experiment & Model & Model Size &  original & ours \\
            \cline{3-5}
      \hline
      \multirow{6}{*}{CIFAR-10}       & \multirow{6}{*}{Clustering} & \multirow{2}{*}{DINOv2}   & base  & $83.63\pm 1.13$ & $\mathbf{84.49\pm 1.19}$ \\
                                      &                             &                           & large & $82.16\pm 1.48$ & $\mathbf{82.49\pm 1.55}$ \\
                                                                    \cline{3-6}
                                      &                             & \multirow{2}{*}{CLIP}     & base  & $72.34\pm 0.80$ & $\mathbf{77.47\pm 1.34}$ \\
                                      &                             &                           & large & $\mathbf{79.45\pm 1.47}$ & $79.25\pm 1.49$ \\
                                                                    \cline{3-6}
                                      &                             & \multirow{2}{*}{DeiT III} & base  & $82.54\pm 3.48$ & $\mathbf{82.89\pm 1.26}$ \\
                                      &                             &                           & large & $84.18\pm 3.11$ & $\mathbf{84.43\pm 2.76}$ \\
      \hline
      \multirow{6}{*}{CIFAR-100}      & \multirow{6}{*}{Clustering} & \multirow{2}{*}{DINOv2}   & base  & $64.26\pm 0.30$ & $\mathbf{65.02\pm 0.14}$ \\
                                      &                             &                           & large & $68.69\pm 0.34$ & $\mathbf{69.04\pm 0.22}$ \\
                                                                    \cline{3-6}
                                      &                             & \multirow{2}{*}{CLIP}     & base  & $42.92\pm 0.22$ & $\mathbf{49.63\pm 0.25}$ \\
                                      &                             &                           & large & $47.88\pm 0.31$ & $\mathbf{56.94\pm 0.21}$ \\
                                                                    \cline{3-6}
                                      &                             & \multirow{2}{*}{DeiT III} & base  & $\mathbf{60.64\pm 0.57}$ & $60.41\pm 0.22$ \\
                                      &                             &                           & large & $\mathbf{67.19\pm 0.57}$ & $67.16\pm 0.59$ \\
      \hline
      \multirow{4}{*}{STL-10}         & \multirow{4}{*}{Clustering} & \multirow{2}{*}{DINOv2}   & base  & $75.65\pm 1.04$ & $\mathbf{82.76\pm 1.27}$ \\
                                      &                             &                           & large & $65.78\pm 1.22$ & $\mathbf{70.51\pm 1.42}$ \\
                                                                    \cline{3-6}
                                      &                             & \multirow{2}{*}{CLIP}     & base  & $85.61\pm 1.74$ & $\mathbf{86.57\pm 1.38}$ \\
                                      &                             &                           & large & $83.67\pm 1.34$ & $\mathbf{84.65\pm 1.39}$ \\
      \hline
      \multirow{4}{*}{Tiny ImageNet}  & \multirow{4}{*}{Clustering} & \multirow{2}{*}{DINOv2}   & base  & $67.81\pm 0.24$ & $\mathbf{68.23\pm 0.25}$ \\
                                      &                             &                           & large & $71.98\pm 0.15$ & $\mathbf{73.19\pm 0.21}$ \\
                                                                    \cline{3-6}
                                      &                             & \multirow{2}{*}{CLIP}     & base  & $35.43\pm 0.16$ & $\mathbf{39.53\pm 0.15}$ \\
                                      &                             &                           & large & $52.54\pm 0.16$ & $\mathbf{55.45\pm 0.14}$ \\
      \hline
      \multirow{6}{*}{CIFAR-100}      & \multirow{6}{*}{k-NN}       & \multirow{2}{*}{DINOv2}   & base  & $87.31$ & $\mathbf{87.58}$ \\
                                      &                             &                           & large & $91.12$ & $\mathbf{91.39}$ \\
                                                                    \cline{3-6}
                                      &                             & \multirow{2}{*}{CLIP}     & base  & $71.72$ & $\mathbf{73.56}$ \\
                                      &                             &                           & large & $78.81$ & $\mathbf{80.90}$ \\
                                                                    \cline{3-6}
                                      &                             & \multirow{2}{*}{DeiT III} & base  & $\mathbf{82.22}$ & $81.97$ \\
                                      &                             &                           & large & $86.13$ & $\mathbf{86.22}$ \\
      \hline
      \multirow{4}{*}{ImageNet-1k}    & \multirow{4}{*}{k-NN}       & \multirow{2}{*}{DINOv2}   & base  & $82.04$ & $\mathbf{82.07}$ \\
                                      &                             &                           & large & $83.50$ & $\mathbf{83.62}$ \\
                                                                    \cline{3-6}
                                      &                             & \multirow{2}{*}{CLIP}     & base  & $73.12$ & $\mathbf{74.26}$ \\
                                      &                             &                           & large & $79.25$ & $\mathbf{80.35}$ \\
      \hline
   \end{tabular}
   }
   \label{tab:models}
\end{table}

\subsection{Comparative Evaluation of Artifacts Attenuation Strategies}
In our proposed method, as described in Sec.~3 of the paper, after the artifacts are identified, the corresponding attention values are attenuated to the minimum value of the patches in each head. Other strategies of attenuation can be considered. In this section, we examine three strategies: (a) Replacing artifacts with $-\infty$ (\textit{-infinity}), (b) replacing artifacts with the average value of attention $\frac{1}{N}\sum_{j}\left((QK^T)_{0j}\right)$ (\textit{average}), (c) replacing artifacts with the minimum value of attention, $\min_{j}\left((QK^T)_{0j}\right)$ (\textit{minimum}), as implemented in our paper. \Cref{tab:attenuating} shows the accuracy of each of the three strategies. \textit{average} outperforms other strategies in STL-10 for ViT-L/14 distilled and ViT-g/14. However, in the cases of CIFAR-100 and Tiny ImageNet for ViT-g/14, it falls below the accuracy of the \textit{original}, and can be considered unstable. The results of the remaining two strategies are not so different, but \textit{minimum} as implemented in our proposed method is slightly more accurate for more cases. There are many other possible variations of the attenuation strategies, but the fact that artifacts attenuation outperforms the original model in accuracy in almost all cases in this study indicates that attenuating attention values is effective to some extent, regardless of the specific value used for substitution.

\begin{table}[t]
   \caption{Image clustering result across various attenuation strategies and model sizes (\textit{small}: ViT-S/14 distilled, \textit{base}: ViT-B/14 distilled, \textit{large}: ViT-L/14 distilled, \textit{giant}: ViT-g/14) reported in ACC.}
   \begin{center}
   \resizebox{\columnwidth}{!}{
   \begin{tabular}{|c|c|c|c|c|c|c|}
   \hline
   Model Size & Method  & CIFAR-10 & CIFAR-100 & STL-10 &Tiny ImageNet\\
   \hline\hline
   \multirow{4}{*}{base}   & original   & $83.63\pm 1.13$ & $64.26\pm 0.30$ & $75.65\pm 1.04$ & $67.81\pm 0.24$ \\
                           & -infinity  & $84.47\pm 1.33$ & $64.92\pm 0.23$ & $82.68\pm 1.25$ & $\mathbf{68.27\pm 0.23}$ \\
                           & average    & $84.27\pm 1.30$ & $64.88\pm 0.33$ & $\mathbf{82.87\pm 1.28}$ & $68.34\pm 0.22$ \\
                           & minimum    & $\mathbf{84.49\pm 1.19}$ & $\mathbf{65.02\pm 0.14}$ & $82.76\pm 1.27$ & $68.23\pm 0.25$ \\
   \hline
   \multirow{4}{*}{large}  & original   & $82.16\pm 1.48$ & $68.69\pm 0.34$ & $65.78\pm 1.22$ & $71.98\pm 0.15$ \\
                           & -infinity  & $\mathbf{82.49\pm 1.56}$ & $69.03\pm 0.21$ & $70.53\pm 1.22$ & $73.18\pm 0.19$ \\
                           & average    & $82.41\pm 1.27$ & $68.75\pm 0.37$ & $\mathbf{72.13\pm 1.39}$ & $72.67\pm 0.18$ \\
                           & minimum    & $\mathbf{82.49\pm 1.55}$ & $\mathbf{69.04\pm 0.22}$ & $70.51\pm 1.42$ & $\mathbf{73.19\pm 0.21}$ \\
   \hline
   \multirow{4}{*}{giant}  & original   & $78.09\pm 1.25$ & $68.99\pm 0.39$ & $55.91\pm 1.14$ & $73.25\pm 0.16$ \\
                           & -infinity  & $78.64\pm 1.87$ & $\mathbf{69.56\pm 0.33}$ & $56.00\pm 0.84$ & $73.52\pm 0.16$ \\
                           & average    & $\mathbf{79.19\pm 1.87}$ & $68.58\pm 0.25$ & $\mathbf{56.25\pm 0.88}$ & $72.97\pm 0.19$ \\
                           & minimum    & $78.59\pm 1.91$ & $69.50\pm 0.28$ & $56.01\pm 0.93$ & $\mathbf{73.54\pm 0.17}$ \\
   \hline
   \end{tabular}
   }
   \end{center}
   \label{tab:attenuating}
\end{table}

\subsection{Comparative Evaluation with LSA}
Locality Self-Attention (LSA) introduced temperature scaling and diagonal masking of the attention matrix to improve local induction bias~\cite{lee2021vision}. We implemented the same diagonal masking of LSA, but only at inference time and only for the final layer of the model to conform with our framework. \Cref{tab:LSA} shows the accuracy of LSA, our method and the original model. When comparing to the original model, LSA is effective for ViT-g/14 and ViT-L/14 distilled, but its accuracy does not clearly improve for ViT-B/14 distilled. LSA also achieves lower accuracy than our method in more cases. In this evaluation, we also investigate the combination of our method with LSA, denoted as \textit{LSA + ours} in \cref{tab:LSA}. The results show that combination with our method improved accuracy in models where LSA was effective. We speculate that while adopting the LSA alone increases the value of artifacts' attention, combining LSA with our method manages to increase the effective attention manipulated by LSA. Because of the complementary relationship between our method and LSA, we believe that the simultaneous adoption of both methods results in the greatest improvement in accuracy.

\begin{table}[t]
   \caption{Image clustering result when adopting LSA in various model sizes (\textit{small}: ViT-S/14 distilled, \textit{base}: ViT-B/14 distilled, \textit{large}: ViT-L/14 distilled, \textit{giant}: ViT-g/14) reported in ACC.}
   \begin{center}
   \resizebox{\columnwidth}{!}{
   \begin{tabular}{|c|c|c|c|c|c|c|}
   \hline
   Model Size & Method  & CIFAR-10 & CIFAR-100 & STL-10 &Tiny ImageNet\\
   \hline\hline
   \multirow{4}{*}{base}      & original   & $83.63\pm 1.13$ & $64.26\pm 0.30$ & $75.65\pm 1.04$ & $67.81\pm 0.24$ \\
                              & LSA        & $83.25\pm 1.41$ & $64.35\pm 0.28$ & $75.77\pm 1.11$ & $67.83\pm 0.15$ \\
                              & ours       & $\mathbf{84.49\pm 1.19}$ & $\mathbf{65.02\pm 0.14}$ & $\mathbf{82.76\pm 1.27}$ & $68.23\pm 0.25$ \\
                              & LSA + ours & $83.97\pm 1.55$ & $64.86\pm 0.37$ & $82.45\pm 1.45$ & $\mathbf{68.33\pm 0.14}$ \\
   \hline
   \multirow{4}{*}{large}  & original   & $82.16\pm 1.48$ & $68.69\pm 0.34$ & $65.78\pm 1.22$ & $71.98\pm 0.15$ \\
                           & LSA        & $82.84\pm 1.68$ & $69.33\pm 0.34$ & $69.39\pm 1.22$ & $72.22\pm 0.20$ \\
                           & ours       & $82.49\pm 1.55$ & $69.04\pm 0.22$ & $70.51\pm 1.42$ & $73.19\pm 0.21$ \\
                           & LSA + ours & $\mathbf{83.14\pm 1.47}$ & $\mathbf{69.75\pm 0.32}$ & $\mathbf{76.58\pm 1.46}$ & $\mathbf{73.60\pm 0.17}$ \\
   \hline
   \multirow{4}{*}{giant}  & original   & $78.09\pm 1.25$ & $68.99\pm 0.39$ & $55.91\pm 1.14$ & $73.25\pm 0.16$ \\
                           & LSA        & $78.82\pm 2.05$ & $69.30\pm 0.33$ & $56.91\pm 0.71$ & $73.36\pm 0.16$  \\
                           & ours       & $78.59\pm 1.91$ & $69.50\pm 0.28$ & $56.01\pm 0.93$ & $73.54\pm 0.17$ \\
                           & LSA + ours & $\mathbf{79.70\pm 1.56}$ & $\mathbf{69.84\pm 0.35}$ & $\mathbf{58.95\pm 0.92}$ & $\mathbf{73.90\pm 0.18}$ \\
   \hline
   \end{tabular}
   }
   \end{center}
   \label{tab:LSA}
\end{table}

\subsection{Comparative Evaluation of Artifacts Identification Strategies}
In Sec.~4 of the paper, we employ the \(L_2\) norms of the query as QKV patches to identify the artifact. However, it is also possible to utilize the \(L_2\) norms of the key or value patches, as described in Sec.~3. Moreover, in prior work, artifacts were identified by the \(L_2\) norms of the final output patch tokens of the model~\cite{darcet2023vision}. In this section, we discuss these other strategies of artifacts identification. The results of clustering are shown in \cref{tab:qkv_norm} after identifying the artifacts by a threshold value $\theta$ for each model from the histogram of \(L_2\) norms of each strategy. From the result, we observe that the method using \textit{output} has a high accuracy in ViT-g/14. However, the clustering accuracy is not stable in ViT-B/14 distilled. As described in Sec.~3, the histogram obtained from the \(L_2\) norms of \textit{output} exhibits unclear bimodality. Hence, there is difficulty in determining an appropriate threshold for identifying artifacts. Also, there is a disadvantage in terms of computational cost when computing artifacts from the \textit{output} as a back pass loop from the output of the model is needed. For comparison between utilizing \textit{query}, \textit{key}, and \textit{value}, we observe that the accuracy of \textit{key}, like the \textit{output}, is not stable for ViT-B/14 distilled, although to a lesser extent. \textit{value} and \textit{query} have almost the same level of accuracy. Therefore, for the sake of clarity, we mainly utilized \textit{query} in our experiments.

\begin{table}[t]
   \caption{Image clustering result across various artifacts identification strategies and model sizes (\textit{small}: ViT-S/14 distilled, \textit{base}: ViT-B/14 distilled, \textit{large}: ViT-L/14 distilled, \textit{giant}: ViT-g/14) reported in ACC.}
   \begin{center}
   \resizebox{\columnwidth}{!}{
   \begin{tabular}{|c|c|c|c|c|c|c|}
   \hline
   Model Size & Method  & CIFAR-10 & CIFAR-100 & STL-10 &Tiny ImageNet\\
   \hline\hline
   \multirow{5}{*}{base}      & original   & $83.63\pm 1.13$ & $64.26\pm 0.30$ & $75.65\pm 1.04$ & $67.81\pm 0.24$ \\
                              & query      & $\mathbf{84.49\pm 1.19}$ & $\mathbf{65.02\pm 0.14}$ & $82.76\pm 1.27$ & $68.23\pm 0.25$ \\
                              & key        & $83.13\pm 1.06$ & $64.78\pm 0.31$ & $79.03\pm 1.22$ & $68.14\pm 0.11$ \\
                              & value      & $84.29\pm 0.98$ & $64.98\pm 0.22$ & $\mathbf{82.77\pm 1.44}$ & $\mathbf{68.30\pm 0.18}$ \\
                              & output     & $\mathbf{84.49\pm 1.67}$ & $64.53\pm 0.29$ & $75.82\pm 1.06$ & $67.96\pm 0.18$ \\
   \hline
   \multirow{5}{*}{large}  & original   & $82.16\pm 1.48$ & $68.69\pm 0.34$ & $65.78\pm 1.22$ & $71.98\pm 0.15$ \\
                           & query      & $82.49\pm 1.55$ & $69.04\pm 0.22$ & $70.51\pm 1.42$ & $73.19\pm 0.21$ \\
                           & key        & $82.92\pm 1.73$ & $69.09\pm 0.32$ & $\mathbf{70.66\pm 1.31}$ & $73.17\pm 0.17$ \\
                           & value      & $82.49\pm 1.56$ & $69.05\pm 0.18$ & $70.56\pm 1.41$ & $\mathbf{73.21\pm 0.16}$ \\
                           & output     & $\mathbf{82.97\pm 1.82}$ & $\mathbf{69.15\pm 0.30}$ & $69.58\pm 1.14$ & $73.03\pm 0.18$ \\
   \hline
   \multirow{5}{*}{giant}  & original   & $78.09\pm 1.25$ & $68.99\pm 0.39$ & $55.91\pm 1.14$ & $73.25\pm 0.16$ \\
                           & query      & $78.59\pm 1.91$ & $69.50\pm 0.28$ & $56.01\pm 0.93$ & $\mathbf{73.54\pm 0.17}$ \\
                           & key        & $78.63\pm 1.88$ & $69.39\pm 0.27$ & $56.01\pm 0.94$ & $73.52\pm 0.20$ \\
                           & value      & $78.59\pm 1.91$ & $69.51\pm 0.30$ & $56.01\pm 0.93$ & $73.53\pm 0.19$ \\
                           & output     & $\mathbf{78.65\pm 1.86}$ & $\mathbf{69.54\pm 0.29}$ & $\mathbf{56.02\pm 0.92}$ & $73.52\pm 0.16$ \\
   \hline
   \end{tabular}
   }
   \end{center}
   \label{tab:qkv_norm}
\end{table}

\subsection{Output Feature Representation Visualization}

\begin{figure}[t]
   \centering
   \begin{tabular}{cc}
      \includegraphics[height=4.5cm]{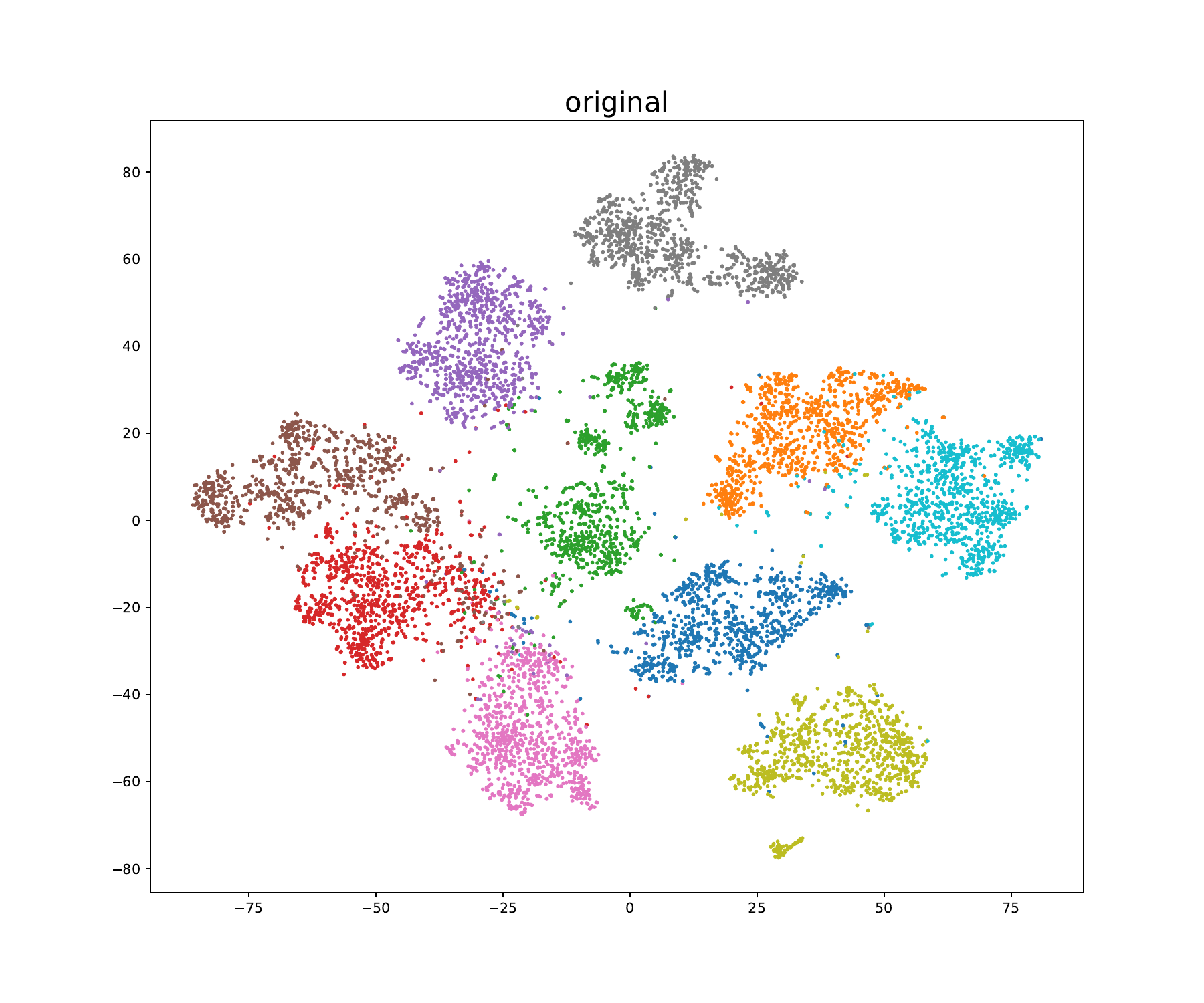}&
      \includegraphics[height=4.5cm]{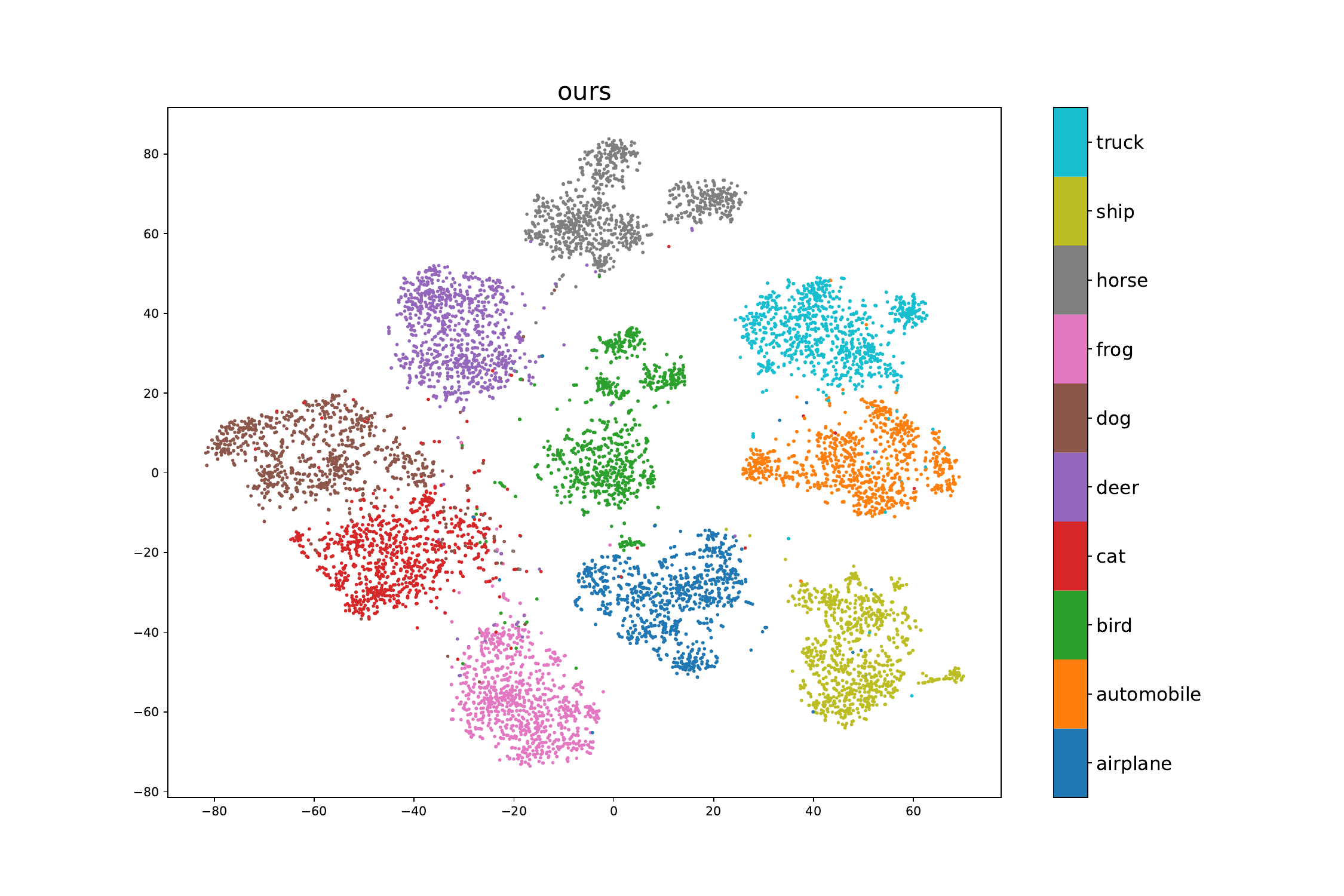}
   \end{tabular}
   \caption{t-SNE visualization: The left map is the output feature map of original model and the right map is the feature map of the model incorporating our proposed method. For our method in this visualization, we set $\theta = 3.5$ as obtained from Sec.~5.2 of the paper for better representing the potential of our method. Colors indicate true labels of image data points. Feature map of our proposed method shows fewer breakaway data points. Best viewed in color (model: ViT-B/14 distilled, dataset: CIFAR-10). }
   \label{fig:TSNE}
\end{figure}

\Cref{fig:TSNE} shows the t-SNE\cite{JMLR:v9:vandermaaten08a} visualization of the output features for the original model and model incorporating our proposed method by using the dataset of CIFAR-10. For the image clustering of CIFAR-10 with ViT-B/14 model, experiment using the original model has a clustering accuracy of 83.63, which improves to 84.49 for $\theta=3.0$ and 84.86 for $\theta=3.5$ by applying our proposed method. It is observed that feature map of our proposed method shows fewer breakaway data points. To quantify this, in the discussion here, we define breakaway points as data points with a silhouette score smaller than 0, calculated using the true labels. For the original model, there are 635 breakaway points out of 10,000 data points. The number of breakaway points decreases to 569 for $\theta=3.0$ and 535 for $\theta=3.5$. These improvements in output features' quality enhance the subsequent image clustering accuracy. 

\section{Limitation}
\subsection{Limitations on Data}
Because this method utilizes a pretrained model and does not involve any re-training, there is a possibility that clustering may not work well on some highly specific datasets due to the bias of the pretrained model. However, for the datasets that we have evaluated, our method proves effective.

\subsection{Limitations on Methodology}
Methods without re-training such as our proposed method maybe difficult to extract performance beyond the potential of the original model. However, algorithms of tuning-free merging of weights from other external models~\cite{xu2024training, huang2024emr} have been proposed. These methods may provide a complimentary solution to our method for better performance.

\section{Licence info}
\label{sec:license}

\Cref{tab:license} shows the license info of images used in Fig.~4 of the paper. The images are overlaid with attention map in the figure.

\begin{table}[h!]
   \caption{License info of images in Fig.~4 of the paper (※urls of number 2 and 9 are currently invalid).}
   \begin{center}
   \resizebox{\columnwidth}{!}{
   \begin{tabular}{|c|c|c|c|}
         \hline
      Number & Image id & URL & License \\
      \hline

      1  & 526751 & \url{ http://farm4.staticflickr.com/3288/2933360267_ae24740821_z.jpg } & Attribution-NonCommercial-NoDerivs License \\
      2 & 574315 & \url{ http://farm3.staticflickr.com/2010/2247055627_5269f84985_z.jpg } & Attribution-NonCommercial-NoDerivs License \\
      3 & 5037 & \url{ http://farm8.staticflickr.com/7379/9599671465_8a2f486da1_z.jpg } & Attribution-NoDerivs License \\
      4 & 246883 & \url{ http://farm4.staticflickr.com/3067/2869541146_a627d12677_z.jpg } & Attribution-NonCommercial-ShareAlike License \\
      5 & 253433 & \url{ http://farm1.staticflickr.com/162/361912851_59c9993d91_z.jpg } & Attribution-NonCommercial-NoDerivs License \\
      6 & 231237 & \url{ http://farm4.staticflickr.com/3607/3342869781_cd4a4b1154_z.jpg } & Attribution-NonCommercial-ShareAlike License \\
      7 & 289659 & \url{ http://farm8.staticflickr.com/7031/6786602747_b7b811b0d5_z.jpg } & Attribution-NonCommercial-NoDerivs License \\
      8 & 163290 & \url{ http://farm7.staticflickr.com/6166/6190069448_f9da6727e6_z.jpg } & Attribution-NonCommercial License \\
      9 & 66817 & \url{ http://farm8.staticflickr.com/7279/7864913910_9e85e0a82a_z.jpg } & Attribution License \\
      10 & 424545 & \url{ http://farm4.staticflickr.com/3193/3054220374_d2a3456295_z.jpg } & Attribution-NonCommercial-NoDerivs License \\
      11 & 378673 & \url{ http://farm8.staticflickr.com/7124/7810951178_b622f1466c_z.jpg } & Attribution-NonCommercial-NoDerivs License \\
      12 & 405279 & \url{ http://farm1.staticflickr.com/236/443807489_3d7fba2557_z.jpg } & Attribution License \\
      13 & 332570 & \url{ http://farm2.staticflickr.com/1051/1392968224_0f863f4054_z.jpg } & Attribution-NonCommercial License \\
      14 & 131386 & \url{ http://farm6.staticflickr.com/5074/5860045248_f99b35c5c8_z.jpg } & Attribution-ShareAlike License \\
      15 & 338560 & \url{ http://farm5.staticflickr.com/4044/4583091116_28eaab2a2b_z.jpg } & Attribution License \\
      \hline
      \multicolumn{4}{l}{Attribution-NonCommercial-ShareAlike License : \url{http://creativecommons.org/licenses/by-nc-sa/2.0/ }} \\
      \multicolumn{4}{l}{Attribution-NonCommercial License : \url{http://creativecommons.org/licenses/by-nc/2.0/ }} \\
      \multicolumn{4}{l}{Attribution-NonCommercial-NoDerivs License : \url{http://creativecommons.org/licenses/by-nc-nd/2.0/ }} \\
      \multicolumn{4}{l}{Attribution License : \url{http://creativecommons.org/licenses/by/2.0/ } }\\
      \multicolumn{4}{l}{Attribution-ShareAlike License : \url{http://creativecommons.org/licenses/by-sa/2.0/ }} \\
      \multicolumn{4}{l}{Attribution-NoDerivs License : \url{http://creativecommons.org/licenses/by-nd/2.0/}} \\
   \end{tabular}
   }
   \end{center}
   \label{tab:license}
\end{table}

\end{document}